\documentclass[conference]{IEEEtran}

\usepackage{cite}
\usepackage{url}
\usepackage{amsmath,amssymb,amsfonts}
\usepackage{algorithm}
\usepackage[noend]{algpseudocode}
\usepackage{graphicx}
\usepackage{subfig}
\usepackage{textcomp}
\usepackage{array}
\usepackage{csquotes}
\usepackage{multirow}
\usepackage{xcolor}
\usepackage{makecell}
\usepackage{subcaption}
\usepackage[hidelinks]{hyperref}
\usepackage[font=footnotesize]{caption}
\usepackage{stfloats}
\usepackage{fancyhdr}
\usepackage{lettrine}
\definecolor{my_blue}{RGB}{0,0,255}

\newcolumntype{C}[1]{>{\centering\arraybackslash}p{#1}}

\begin{document}

\title{DeepDetect: Learning All-in-One Dense Keypoints}

\author{
\hspace{0.0cm}\IEEEauthorblockN{Shaharyar Ahmed Khan Tareen}
\hspace{0.0cm}\textit{University of Houston}\\
\hspace{0.0cm}Houston, TX, USA \\
\hspace{0.0cm}stareen@cougarnet.uh.edu
\and
\IEEEauthorblockN{Filza Khan Tareen}\hspace{0.2cm}
\textit{National University of Sciences and Technology}\hspace{0.6cm} \\
Islamabad, Pakistan \hspace{-0.23cm} \\ ftareen.bse2021mcs@student.nust.edu.pk
\hspace{-0.05cm} \\
\and
\hspace{2.2cm}
\IEEEauthorblockN{\phantom{Lei Fan}}
\hspace{2.0cm}\textit{\phantom{University of Houston}}\\
\hspace{2.2cm}\phantom{Houston, TX, USA} \\
\hspace{2.2cm}\phantom{lfan8@central.uh.edu}

\and
\IEEEauthorblockN{Xiaojing Yuan}
\hspace{-0.2cm}\textit{University of Houston}\\
\hspace{0.0cm}Houston, TX, USA \\
\hspace{0.0cm}xyuan@uh.edu

\and
\IEEEauthorblockN{\phantom{Xiaojing Yuan}}
\hspace{-0.2cm}\textit{\phantom{University of Houston}}\\
\hspace{0.0cm}\phantom{Houston, TX, USA} \\
\hspace{0.0cm}\phantom{xyuan@uh.edu}
}

\maketitle

\thispagestyle{fancy}
\fancyhf{}
\fancyfoot[C]{\textbf{Preprint:} Accepted at the 2026 IEEE International Conference on AI and Data Analytics
(ICAD 2026), Boston, MA, USA}
\renewcommand{\headrulewidth}{0pt}
\renewcommand{\footrulewidth}{0pt}

\begin{abstract}
Keypoint detection is the foundation of many computer vision tasks, including image registration, structure-from-motion, 3D reconstruction, visual odometry, and SLAM. Traditional detectors (SIFT, ORB, BRISK, FAST, etc.) and learning-based methods (SuperPoint, R2D2, QuadNet, LIFT, etc.) have shown strong performance gains yet suffer from key limitations: sensitivity to photometric changes, low keypoint density and repeatability, limited adaptability to challenging scenes, and lack of semantic understanding, often failing to prioritize visually important regions. We present DeepDetect, an intelligent, all-in-one, dense detector that unifies the strengths of classical detectors using deep learning. Firstly, we create ground-truth masks by fusing outputs of 7 keypoint and 2 edge detectors, extracting diverse visual cues from corners and blobs to prominent edges and textures in the images. Afterwards, a lightweight and efficient model: \enquote{ESPNet}, is trained using fused masks as labels, enabling DeepDetect to focus semantically on images while producing highly dense keypoints, that are adaptable to diverse and visually degraded conditions. Evaluations on Oxford, HPatches, and Middlebury datasets demonstrate that DeepDetect surpasses other detectors achieving maximum values of 0.5143 (average keypoint density), 0.9582 (average repeatability), 338,118 (correct matches), and 842,045 (voxels in stereo 3D reconstruction).
\end{abstract}
\vspace{0.15cm}
\begin{IEEEkeywords}
keypoint detection, edge detection, image fusion, image matching, SIFT, ORB, deep learning, 3D reconstruction.
\vspace{-0.10cm}
\end{IEEEkeywords}

\section{Introduction}
Keypoint detection is the backbone of many computer vision applications including image matching, panorama stitching, object tracking, 3D reconstruction, visual odometry, and visual SLAM \cite{ref_1}. Keypoints (often referred to as features or feature points) are the distinctive points in an image that can also be identified to an extent under various transformations or degradations by the detectors. They are usually corners, blobs, ridges, and textured patterns that stand out from their surroundings \cite{ref_1}. Selection of keypoint detector is a critical decision in computer vision as it determines the stability, repeatability, map density, and overall performance of the downstream application \cite{ref_2,ref_3,ref_4,ref_5,ref_6}. A variety of detectors have been proposed over the years, ranging from traditional algorithms (SIFT, SURF, Harris Corner Detector, ORB, BRISK, and AGAST) to deep learning based approaches (SuperPoint, R2D2, D2-Net, Quad-Net, and LIFT). After keypoint detection, corresponding feature descriptors are computed by encoding the local pixel neighborhood around each keypoint into a distinctive vector that enhances stability across image variations \cite{ref_1}. A good descriptor is more distinctive and invariant to diverse image transformations or degradations \cite{ref_7,ref_8,ref_9,ref_10}.

\begin{figure}[!t]
\vspace{0.12cm}
\centering
\includegraphics[width=1.0\columnwidth]{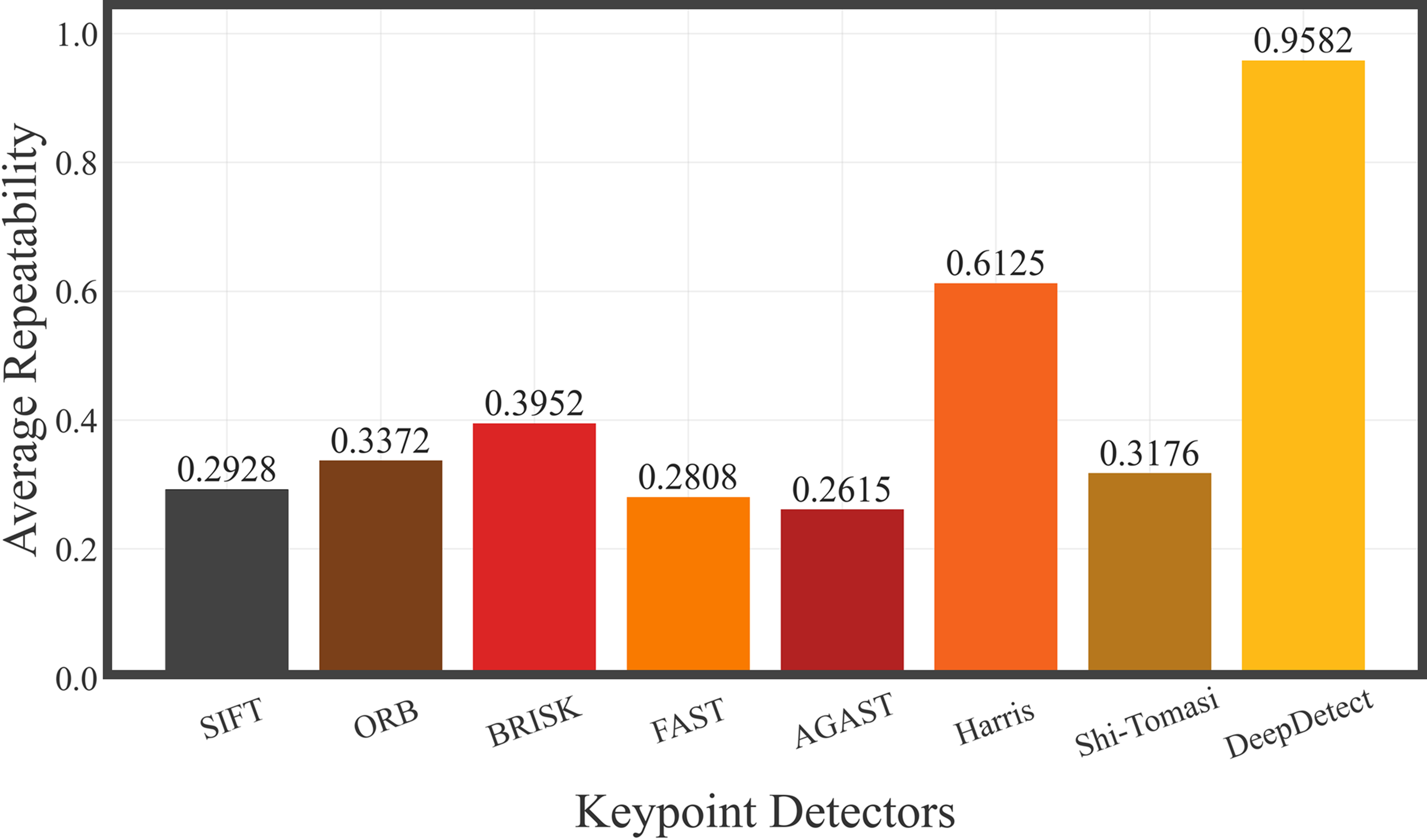}
\caption{\centering Average repeatability of keypoint detectors on Oxford dataset \cite{ref_30}.}
\vspace{-0.50cm}
\label{fig:figure_1}
\end{figure}

\vspace{0.20cm}
After detection and description, \enquote{feature matching} is performed to match the descriptors and establish geometric relationships between the images \cite{ref_7}. NNDR is an effective matching strategy \cite{ref_11}, that combines nearest neighbor search with a distance ratio test to reduce incorrect matches. The resulting matches are filtered by using model fitting algorithms: RANSAC \cite{ref_12}, PROSAC \cite{ref_13}, and MSAC to reject the outliers. Reliable image matching depends on the effectiveness of four stages: keypoint detection, keypoint description, feature (descriptor) matching, and outlier rejection \cite{ref_7}, enabling robots or machines to accurately perceive, localize, and generate dense maps of their surroundings. Therefore, achieving high performance in the primary stage: \textbf{keypoint detection}, is crucial, as weaknesses at this stage propagate and lead to degradation in the performance of downstream tasks.

\begin{figure*}[t]
	\centering
	\begin{minipage}[t]{0.80\textwidth}
    \centering
    \subfloat[Original Image]{
        \begin{minipage}[t]{0.23\textwidth}
            \centering
            \includegraphics[width=\linewidth]{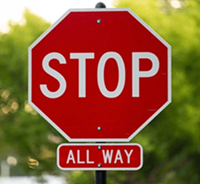}\\[7pt]
            \includegraphics[width=\linewidth]{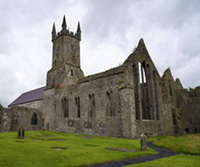}
        \end{minipage}
    }\hfill
    \subfloat[SIFT$_{\text{(Normal)}}$]{
        \begin{minipage}[t]{0.23\textwidth}
            \centering
            \includegraphics[width=\linewidth]{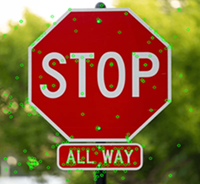}\\[7pt]
            \includegraphics[width=\linewidth]{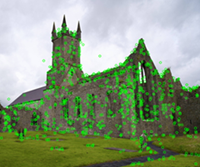}
        \end{minipage}
    }\hfill
    \subfloat[SIFT$_{\text{(Low)}}$]{
        \begin{minipage}[t]{0.23\textwidth}
            \centering
            \includegraphics[width=\linewidth]{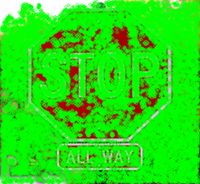}\\[7pt]
            \includegraphics[width=\linewidth]{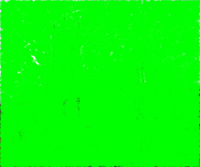}
        \end{minipage}
    }\hfill
    \subfloat[DeepDetect]{
        \begin{minipage}[t]{0.23\textwidth}
            \centering
            \includegraphics[width=\linewidth]{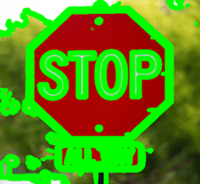}\\[7pt]
            \includegraphics[width=\linewidth]{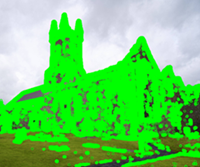}
        \end{minipage}
    }
    \end{minipage}
    \caption{\centering Keypoint detection on two images using SIFT with normal (default) thresholds, SIFT with extremely low thresholds, and DeepDetect. Number of detected keypoints: Top Scene --- 279, 23990, 46309 and Bottom Scene --- 1009, 75465, 51115, respectively. SIFT with standard thresholds yields limited keypoints, whereas SIFT with extremely low thresholds produces high number of keypoints, detecting noisy features on semantically irrelevant regions. DeepDetect produces highly dense keypoints with substantially lower noise, well concentrated in semantically important regions of the images.}
    \vspace{-0.20cm}
	\label{fig:fig_2}
\end{figure*}

\vspace{0.25cm}
Classical detectors are invariant to different image transformations (scale, rotation, affine, blur) but sensitive to extreme variations (intense darkness, extreme photometric/geometric changes, severe smoke or fog) \cite{ref_8}. With standard settings, they detect sparse keypoints that inadequately cover texture-rich and low-contrast regions \cite{ref_8}. They also lack semantic understanding to prioritize important regions in the images. Learning-based detectors SuperPoint, R2D2, D2-Net, Quad-Net, and LIFT) improve keypoint generalization and stability but their qualitative results show similar shortcomings \cite{ref_22,ref_23,ref_25,ref_35,ref_36}. They detect sparse keypoints and lack semantic awareness, potentially missing important regions in cluttered or low-visibility scenes \cite{ref_22,ref_23,ref_25,ref_35,ref_36}. This highlights the need for an intelligent and dense keypoint detector that is robust to complex image transformations and focuses on important regions.

\vspace{0.20cm}
\noindent \textbf{Key Contributions:} Our contributions are highlighted below:
\begin{itemize}
\item We present \textbf{DeepDetect}, an \textbf{intelligent}, \textbf{adaptable}, \textbf{all-in-one}, and \textbf{dense} keypoint detection approach that uses deep learning to learn the strengths of 7 strong keypoint detectors (SIFT, ORB, BRISK, FAST, AGAST, Harris Corner Detector, and Shi-Tomasi Corners) and 2 well-known edge detectors (Canny and Sobel). It provides highly dense and semantically meaningful keypoints, demonstrating robustness to poor visibility and brightness, low contrast, and complex scenes, without requiring manual tuning (as in classical detectors).
\item We name this novel concept of \textbf{keypoint-edge fusion} as \textbf{DeepDetect}, that generates rich and diverse supervision masks, which later on serve as labels for training.
\item We use a lightweight, efficient model: \enquote{ESPNet}, to ensure that DeepDetect has a small memory footprint: \textbf{~1.82 MB} and \textbf{~0.41 M parameters}, making it well-suited for deployment on edge devices such as mobile phones.
\item We demonstrate that DeepDetect outperforms other detectors both qualitatively and quantitatively by significant margins. DeepDetect provides highest performance across three different datasets, with an average repeatability of \textbf{0.9582} (Oxford), average keypoint density of \textbf{0.5143} (Oxford), \textbf{338,118} number of correct matches (HPatches), and as high as \textbf{842,045} voxels in 3D reconstruction (Middlebury), outperforming the classical detectors even when extremely low thresholds are used.
\item The code of DeepDetect is available at the following link: \href{https://github.com/saktx/DeepDetect}{\textbf{https://github.com/saktx/DeepDetect}}
\end{itemize}

\begin{figure*}[!t]
	\centering
    \includegraphics[width=1.0\textwidth]{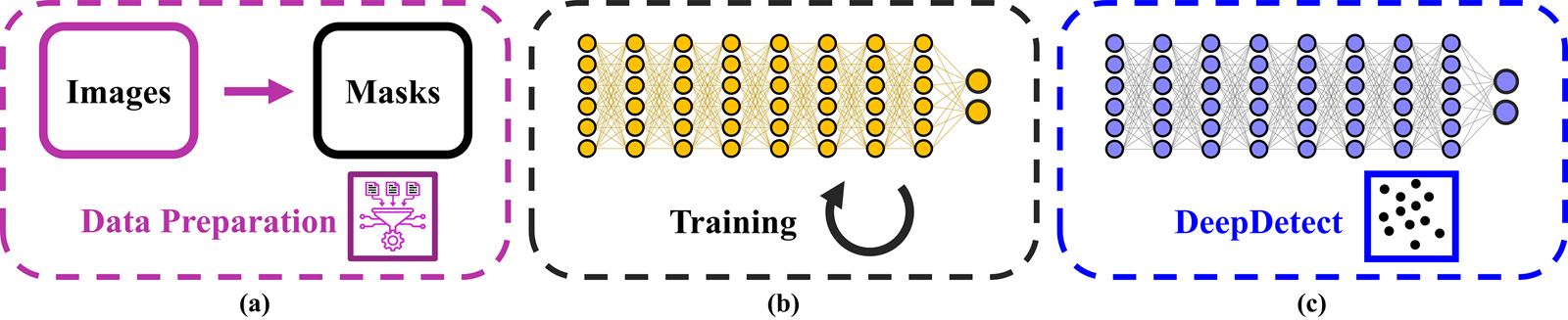}
    \caption{\centering Main stages in the development of DeepDetect: (a) Fusion masks are created by using 7 keypoint detectors (SIFT, ORB, BRISK, FAST, AGAST, Harris, Shi-Tomasi) and 2 edge detectors (Canny, Sobel) in the images, combining their strengths to capture rich and diverse visual representations. (b) The fused binary masks are then used as labels for the corresponding images to supervise and train ESPNet (a lightweight, efficient model). (c) DeepDetect is ready to detect all-in-one, dense keypoints with semantic awareness and adaptability to complex image degradations, without requiring manual tuning.}
    \vspace{-0.15cm}
    \label{fig:fig_3}
\end{figure*}

\section{Related Work}
Classical feature detectors have laid the foundation of modern computer vision. However, they rely on handcrafted criteria such as gradient magnitude, intensity variation, or corner response thresholds. Although computationally efficient, they often fail with standard thresholds under extreme image transformations, lacking adaptability to changing visibility conditions \cite{ref_8}. Learning based methods aim to learn keypoint detection, description (or sometimes both) but they typically produce sparse keypoints \cite{ref_22,ref_23,ref_25,ref_35,ref_36}. which are not suitable for computer vision applications that require dense keypoints such as dense 3D reconstruction, AR/VR, and dense Simultaneous Localization and Mapping (SLAM).

\subsection{Classical Detectors and Descriptors}
SIFT \cite{ref_11}, a powerful scale and rotation invariant algorithm, uses Difference-of-Gaussians for keypoint detection and describes them with gradient-based histograms. SURF \cite{ref_14} is a faster alternative to SIFT that uses integral images and box filters for detection. It also provides feature descriptors but they are less distinctive than SIFT. KAZE \cite{ref_15} detects nonlinear scale-space keypoints directly in the image space yielding highly distinct features but at a higher computational cost whereas AKAZE is accelerated version of KAZE that uses binary descriptors (M-LDB) for faster computation while retaining much of the distinctiveness of KAZE. ORB \cite{ref_16} combines FAST detector with a rotated version of the BRIEF descriptor for speed and rotation invariance.

\vspace{0.20cm}
BRISK \cite{ref_17} uses a circular sampling pattern for detection and generates binary descriptors, offering scale and rotation invariance with high speed and moderate distinctiveness. FAST \cite{ref_18} is an efficient corner detector that examines a circle of 16 pixels around a candidate keypoint. AGAST \cite{ref_19} is an improvement over FAST, designed to be faster and more robust. Harris Corner Detector (HCD) detects corners by analyzing local intensity changes and Shi-Tomasi Corners focus on stronger minimum eigenvalues, leading to more stable corner detection. Canny \cite{ref_20} is a multi-stage detector that uses gradient computation, non-maximum suppression, and hysteresis thresholding to produce thin and fine edges. Sobel \cite{ref_21} uses convolution with gradient kernels to find edges in the x and y directions, providing thicker edges.

\subsection{Learning based Detection and Description}
SuperPoint \cite{ref_22} is a well-known learning based detector and descriptor trained via synthetic homographies, offering high repeatability and descriptor quality across varying conditions. R2D2 \cite{ref_23} learns both repeatability and descriptor reliability via deep learning, filtering out unstable keypoints for robust matching. D2-Net \cite{ref_25} uses a CNN to jointly detect and describe dense keypoints in a single forward pass, particularly effective for challenging photometric and geometric changes. QuadNet is a learning-based detector that uses a quadruplet CNN to learn repeatable and discriminative keypoints without relying on handcrafted heuristics \cite{ref_35}. LIFT \cite{ref_36} learns detection, orientation estimation, and description using three CNNs. Although having some strengths, these algorithms lack semantic awareness about the scenes and do not provide dense keypoints that are especially required in mapping applications such as dense 3D reconstruction, AR/VR, and visual SLAM.

\begin{figure*}[b]
	\centering
	\vspace{-0.35cm}
	\begin{minipage}[t]{1.0\textwidth}
    \centering
    \subfloat[Original Image]{
        \includegraphics[width=0.23\textwidth]{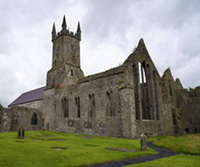}
    }\hfill
    \subfloat[Keypoints Mask]{
        \includegraphics[width=0.23\textwidth]{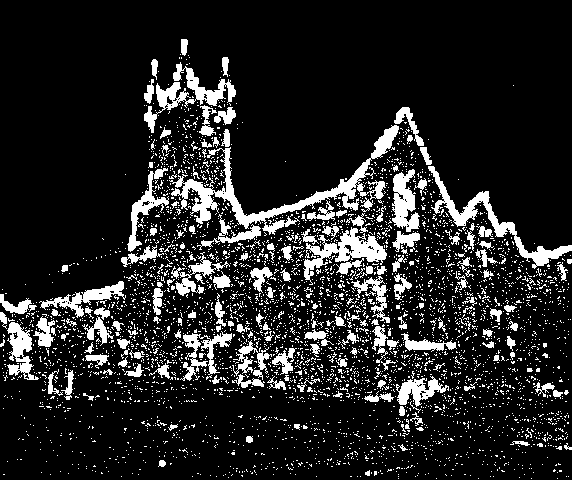}
    }\hfill
    \subfloat[Edges Mask]{
        \includegraphics[width=0.23\textwidth]{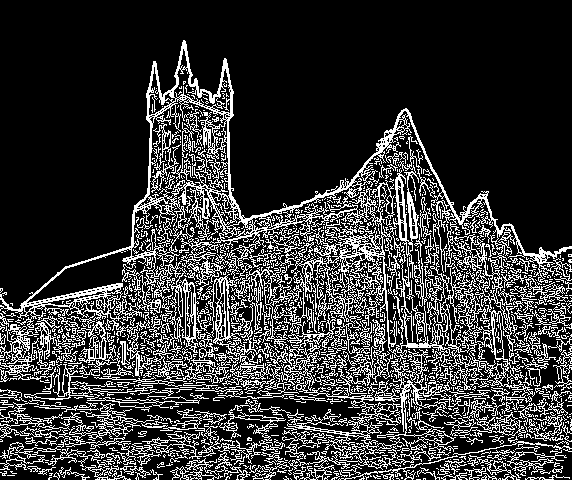}
    }\hfill
    \subfloat[Combined Mask]{
        \includegraphics[width=0.23\textwidth]{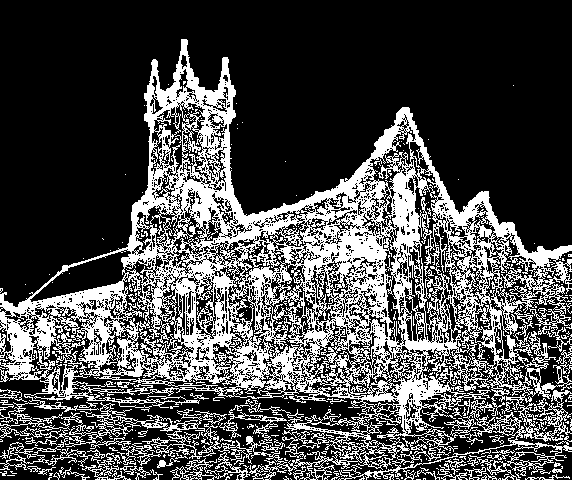}
    }
    \end{minipage}
    \caption{\centering Binary masks obtained from 7 keypoint and 2 edge detectors, along with their combined version (which provides richer representations for training).}
    \vspace{-0.35cm}
    \label{fig:fig_4}
\end{figure*}

\section{DeepDetect -- Methodology}
\subsection{Data Preparation -- Keypoint-Edge Fusion}
Data preparation is the first stage towards the development of DeepDetect as shown in Figure~\ref{fig:fig_3}.
We selected 41000 images from the MS-COCO \cite{ref_26} and NewTsukuba \cite{ref_27} datasets, applying extreme reductions in brightness and contrast to 25\% of them. Afterwards, they were split into training and validation sets of 33000 and 8000 images, respectively. The strengths of the 7 keypoint detectors and 2 edge detectors are then unified by creating a combined binary mask of their outputs for each image. SIFT detects blob-like structures, while the other 6 keypoint detectors identify corners or ridges. Canny highlights fine details, whereas Sobel captures thicker, high-intensity edges in the images. Binary masks are generated with pixels belonging to the detected keypoints-edges marked as 1 and all other pixels as 0. Extremely low detection thresholds are used for the low visibility scenes, whereas moderate thresholds are used for the normal scenes. This mitigates irrelevant keypoints from the normal scenes while maximizing detection in challenging scenes. For each image, its keypoint and edge masks are merged to create a rich, multi-cue supervision mask that serves as its label during training. Given an image \( I \in \mathbb{R}^{H \times W \times 3} \), set of keypoint detectors D = \{SIFT, ORB, BRISK, FAST, AGAST, Harris, Shi-Tomasi\} and set of edge detectors E = \{Canny, Sobel\}, the final combined mask M(I), shown in Figure~\ref{fig:fig_4}(d), is obtained as:

\begin{equation}
M(I) = \bigvee_{d \in D} M_D(I) \,\vee\, \bigvee_{e \in E} M_E(I)
\label{eq:eq_1}
\end{equation}

where $\vee$ denotes the pixel wise logical OR operation, $M_{D}(I)$ shows the fusion mask obtained from the keypoint detectors as in Figure~\ref{fig:fig_4}(b) and $M_{E}(I)$ shows the fusion mask obtained from the edge detectors as in Figure~\ref{fig:fig_4}(c).



\subsection{Learning All-in-One Keypoints -- Training Stage}
ESPNet \cite{ref_31} is a light weight model designed for highly efficient pixel-wise prediction tasks such as \enquote{semantic segmentation} on resource-constrained devices. It has encoder-decoder architecture: the encoder captures contextual information through successive convolution and down-sampling layers while the decoder progressively up-samples the feature maps to restore the spatial resolution. ESPNet is based on ESP modules which factorize the standard convolutions into point-wise $(1\times1)$ projections followed by spatial pyramids of dilated convolutions \cite{ref_31}. ESPNet works on a lower dimensional space due to the point-wise projections, reducing its memory footprint and computational cost. Compared to heavy image segmentation models like U-Net \cite{ref_28} and PSPNet \cite{ref_32}, it is upto $180\times$ smaller, $22\times$ faster, and an excellent choice for low complexity tasks \cite{ref_31}.

\begin{figure}[!t]
	\centering
	\vspace{0.07cm}
    \includegraphics[width=0.62\columnwidth]{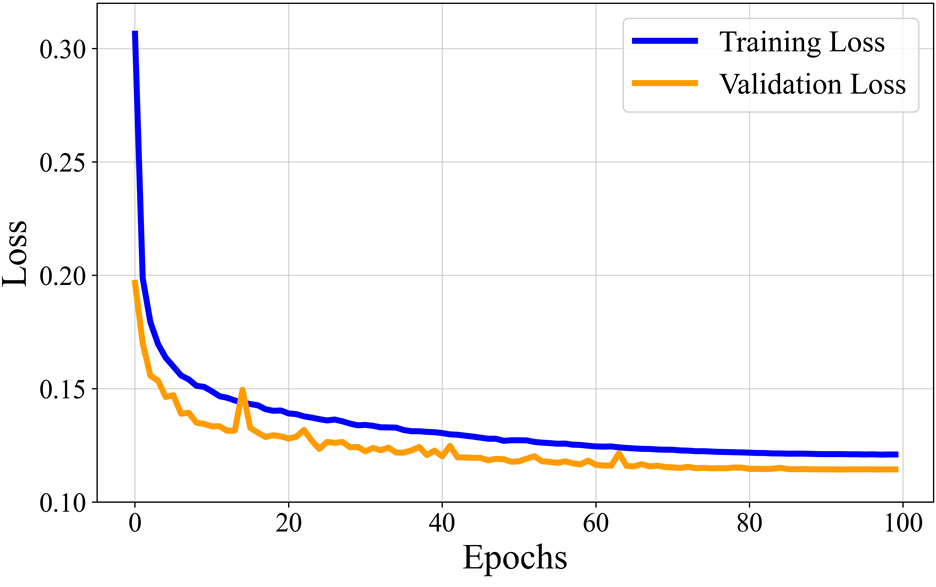}
    \caption{\centering Training and validation loss curves of DeepDetect.}
    \vspace{-0.40cm}
    \label{fig:fig_5}
\end{figure}

\vspace{0.20cm}
In the training stage (Figure~\ref{fig:fig_3}), ESPNet is trained on the prepared dataset for $100$ epochs with a learning rate of $0.001$ and batch size of $64$, using cosine annealing scheduler for smooth convergence. We used Python, OpenCV, and PyTorch \cite{ref_29} for coding. The input image size is kept consistent at $3\times480\times480$ for the model and its predicted output mask is re-sized to match the spatial dimensions of the original image. Given a target mask $M(I) = y \in \{0, 1\}^{H \times W}$ and model's raw output $z \in \mathbb{R}^{H \times W}$, the BCE loss $L(y, z)$ is described as:
\begin{equation}
L = -\frac{1}{N} \sum_{i=1}^{N} \Big[ y_i \log (\sigma(z_i)) + (1-y_i) \log (1-\sigma(z_i)) \Big]
\label{eq:eq_3}
\end{equation}

where $\sigma$ is the sigmoid function and $N = H \times W$ represents the total number of pixels in the mask. The minimization of loss in Equation~\ref{eq:eq_3} encourages the model to produce probabilities closer to $1$ for the keypoint pixels and closer to $0$ for others. Figure~\ref{fig:fig_5} shows the training and validation loss curves of DeepDetect. We picked the model that provided best validation loss during training. During inference, the output $z$ of the model is passed through the sigmoid function to generate the probability $P_p = \sigma(z)$ for pixel $p$. The predicted output of the model is then thresholded at $\tau = 0.5$ to obtain the final binary mask. The threshold can be adjusted using Equation~\ref{eq:eq_4} as per the required keypoint density in the images.

\begin{equation}
M(I) = 
\begin{cases}
1, & \text{if } P_p \ge \tau \\
0, & \text{otherwise}
\end{cases}
\label{eq:eq_4}
\end{equation}

\begin{table}[b]
\definecolor{my_blue}{RGB}{0,0,255}
\definecolor{my_green}{RGB}{29,153,66}
\definecolor{my_red}{RGB}{255,0,0}
\renewcommand{\arraystretch}{1.1}
\vspace{-0.15cm}
\centering
\scriptsize
\caption{\centering Comparison of Average Keypoint Density and Foreground-to-Keypoint (F-KP) Ratio of the Detectors.}
\label{tab:tab_1}
\begin{tabular}{lcc}
\hline
\makecell[c]{\small \textbf{Detector}} & 
\makecell[c]{\footnotesize \textbf{Average} \\ \footnotesize \textbf{Keypoint Density}} & 
\makecell[c]{\footnotesize \textbf{F-KP Ratio}} \\
\hline
SIFT                     & 0.0090 & \textcolor{my_green}{\textbf{0.7606}} \\
SIFT (Low)               & 0.1993 & \textcolor{my_red}{\textbf{0.3559}} \\
ORB                      & 0.0009 & \textcolor{my_green}{\textbf{0.8508}} \\
ORB (Low)                & 0.1256 & \textcolor{my_green}{\textbf{0.6797}} \\
BRISK                    & 0.0127 & \textcolor{my_green}{\textbf{0.7909}} \\
BRISK (Low)              & 0.0800 & \textcolor{my_red}{\textbf{0.5716}} \\
FAST                     & 0.0270 & \textcolor{my_green}{\textbf{0.7442}} \\
FAST (Low)               & 0.3664 & \textcolor{my_red}{\textbf{0.5387}} \\
AGAST                    & 0.0280 & \textcolor{my_green}{\textbf{0.7363}} \\
AGAST (Low)              & 0.3550 & \textcolor{my_red}{\textbf{0.5669}} \\
Harris                   & 0.0246 & \textcolor{my_green}{\textbf{0.8262}} \\
Harris (Low)             & 0.2988 & \textcolor{my_green}{\textbf{0.7389}} \\
Shi-Tomasi               & 0.0052 & \textcolor{my_green}{\textbf{0.8026}} \\
Shi-Tomasi (Low)         & 0.0326 & \textcolor{my_red}{\textbf{0.6180}} \\
\hline
DeepDetect ($\tau = 0.9$) & \textcolor{my_blue}{\textbf{0.4170}} & \textcolor{my_green}{\textbf{0.7114}} \\
DeepDetect ($\tau = 0.8$) & \textcolor{my_blue}{\textbf{0.4493}} & \textcolor{my_green}{\textbf{0.7113}} \\
DeepDetect ($\tau = 0.7$) & \textcolor{my_blue}{\textbf{0.4733}} & \textcolor{my_green}{\textbf{0.7109}} \\
DeepDetect ($\tau = 0.5$) & \textcolor{my_blue}{\textbf{0.5143}} & \textcolor{my_green}{\textbf{0.7093}} \\
\hline
\end{tabular}
\end{table}

\subsection{DeepDetect --- Dense Keypoint Detection}
DeepDetect is not only learned to reproduce the geometric diversity of the classical keypoint and edge detectors but also able to capture semantically important regions, leading to dense and foreground focused keypoints. To perform image matching, keypoints are first detected in the image pair using the trained ESPNet. Afterwards, we use SIFT descriptor to compute the corresponding descriptor vectors, because of its highly distinctive nature. The descriptor sets for the two images are then matched using NNDR based matching. RANSAC can be applied to reject the outliers and filter the correct matches. Qualitative results of DeepDetect are shown in Figure~\ref{fig:fig_6}. For the comparative analysis presented in Table~\ref{tab:tab_2} and Figure~\ref{fig:fig_6}, we used the ground-truth Homography matrices \cite{ref_30} (not RANSAC) to obtain the number of correct matches.

\section{Experiments \& Results}
We compare DeepDetect with other keypoint detectors on the Oxford Affine Covariant Regions Dataset \cite{ref_30} using the following evaluation metrics. In all experiments, we used the SIFT descriptor with all the detectors except ORB and BRISK, for which we used their own descriptors.
\vspace{0.20cm}

\textbf{Keypoint Density:} Shows how many keypoints are detected per unit image area. It helps assess whether a detector produces sparse or dense keypoints. Given $N$ detected keypoints in an image with area $H \times W$, the keypoint density is calculated as:
\begin{equation}
\text{Keypoint Density} = \frac{N}{H \times W}
\label{eq:eq_5}
\end{equation}

\textbf{Repeatability:} Measures the consistency of a detector in finding the same keypoints in images under a given transformation. If $N_A$ and $N_B$ are the number of keypoints detected in two overlapping images $A$ and $B$, and $N_{A \cap B}$ is the number of keypoints within the common region $A \cap B$, then:
\begin{equation}
\text{Repeatability} = \frac{N_{A \cap B}}{\min(N_A, N_B)}
\label{eq:eq_6}
\end{equation}

\begin{figure*}[!t]
	\centering
        \begin{minipage}[t]{0.244\textwidth}
            \centering
            \includegraphics[width=\linewidth]{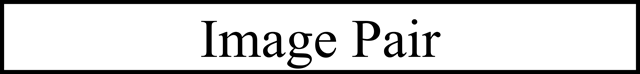}
            \\[3pt]
            \includegraphics[width=\linewidth]{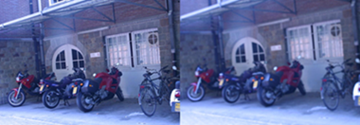}\\[-5pt]
            \caption*{\centering Bikes-1 and Bikes-6 \cite{ref_30}}
            \includegraphics[width=\linewidth]{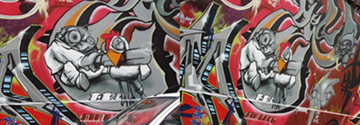}\\[-5pt]
            \caption*{\centering Graf-1 and Graf-3 \cite{ref_30}}
            \includegraphics[width=\linewidth]{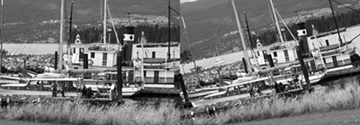}\\[-5pt]
            \caption*{\centering Boat-1 and Boat-2 \cite{ref_30}}
            \includegraphics[width=\linewidth]{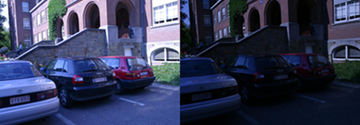}\\[-5pt]
            \caption*{\centering Leuven-1 and Leuven-6 \cite{ref_30}}
            \includegraphics[width=\linewidth]{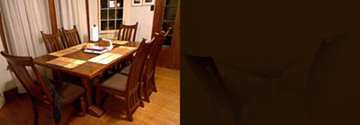}\\[-5pt]
            \caption*{\centering Dusty-1 and Dusty-2 \cite{ref_8}}
            \includegraphics[width=\linewidth]{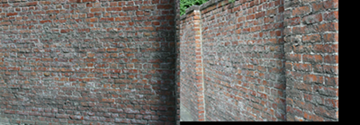}\\[-5pt]
            \caption*{\centering Wall-1 and Wall-6 \cite{ref_30}}
            \includegraphics[width=\linewidth]{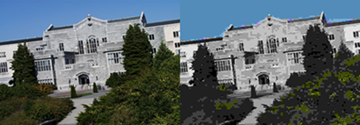}\\[-5pt]
            \caption*{\centering UBC-1 and UBC-6 \cite{ref_30}}
            \includegraphics[width=\linewidth]{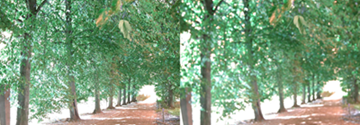}\\[-5pt]
            \caption*{\centering Trees-1 and Trees-6 \cite{ref_30}}
        \end{minipage}
        \begin{minipage}[t]{0.244\textwidth}
            \centering
            \includegraphics[width=\linewidth]{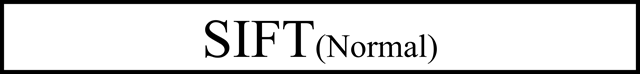}\\[3pt]
            \includegraphics[width=\linewidth]{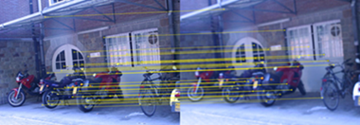}\\[-5pt]
            \caption*{\centering 50 Correct Matches}
            \includegraphics[width=\linewidth]{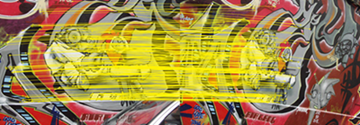}\\[-5pt]
            \caption*{\centering 368 Correct Matches}
            \includegraphics[width=\linewidth]{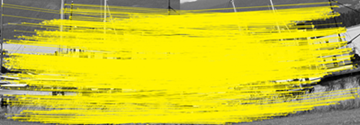}\\[-5pt]
            \caption*{\centering 2,202 Correct Matches}
            \includegraphics[width=\linewidth]{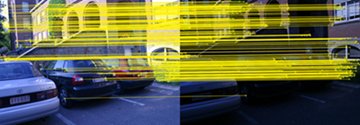}\\[-5pt]
            \caption*{\centering 355 Correct Matches}
            \includegraphics[width=\linewidth]{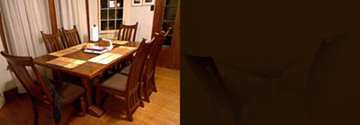}\\[-5pt]
            \caption*{\centering 0 Correct Matches}
            \includegraphics[width=\linewidth]{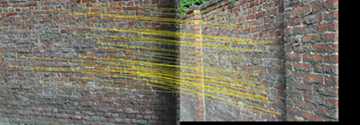}\\[-5pt]
            \caption*{\centering 55 Correct Matches}
            \includegraphics[width=\linewidth]{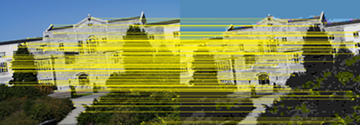}\\[-5pt]
            \caption*{\centering 252 Correct Matches}
            \includegraphics[width=\linewidth]{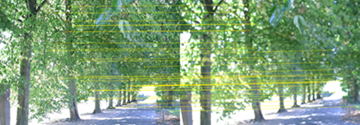}\\[-5pt]
            \caption*{\centering 39 Correct Matches}
        \end{minipage}
        \begin{minipage}[t]{0.244\textwidth}
            \centering
            \includegraphics[width=\linewidth]{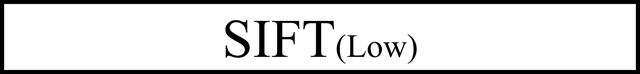}\\[3pt]
            \includegraphics[width=\linewidth]{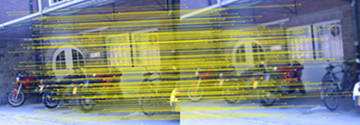}\\[-5pt]
            \caption*{\centering 268 Correct Matches}
            \includegraphics[width=\linewidth]{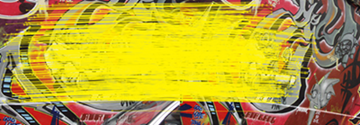}\\[-5pt]
            \caption*{\centering 1,020 Correct Matches}
            \includegraphics[width=\linewidth]{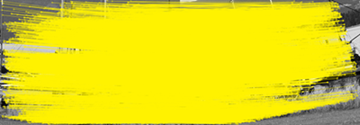}\\[-5pt]
            \caption*{\centering 6,517 Correct Matches}
            \includegraphics[width=\linewidth]{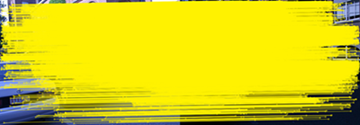}\\[-5pt]
            \caption*{\centering 7,822 Correct Matches}
            \includegraphics[width=\linewidth]{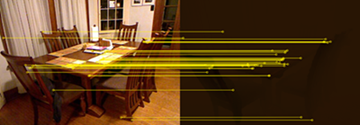}\\[-5pt]
            \caption*{\centering 45 Correct Matches}
            \includegraphics[width=\linewidth]{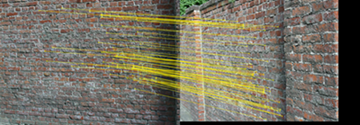}\\[-5pt]
            \caption*{\centering 66 Correct Matches}
            \includegraphics[width=\linewidth]{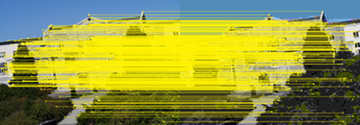}\\[-5pt]
            \caption*{\centering 790 Correct Matches}
            \includegraphics[width=\linewidth]{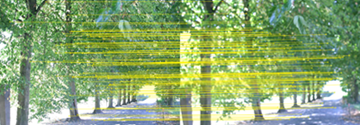}\\[-5pt]
            \caption*{\centering 134 Correct Matches}
        \end{minipage}
        \begin{minipage}[t]{0.244\textwidth}
            \centering
            \includegraphics[width=\linewidth]{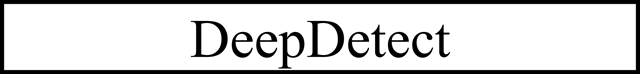}\\[3pt]
            \includegraphics[width=\linewidth]{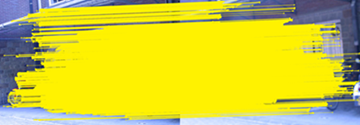}\\[-5pt]
            \caption*{\centering 16,581 Correct Matches}
            \includegraphics[width=\linewidth]{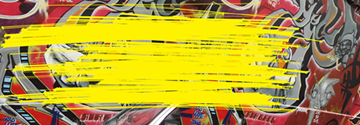}\\[-5pt]
            \caption*{\centering 2,715 Correct Matches}
            \includegraphics[width=\linewidth]{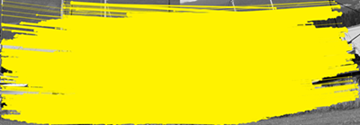}\\[-5pt]
            \caption*{\centering 47,447 Correct Matches}
            \includegraphics[width=\linewidth]{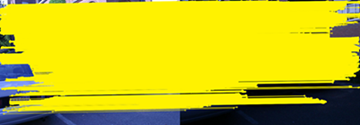}\\[-5pt]
            \caption*{\centering 59,003 Correct Matches}
            \includegraphics[width=\linewidth]{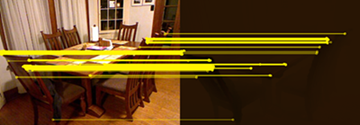}\\[-5pt]
            \caption*{\centering 502 Correct Matches}
            \includegraphics[width=\linewidth]{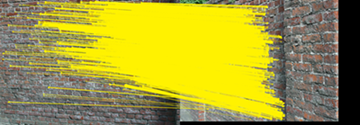}\\[-5pt]
            \caption*{\centering 4,756 Correct Matches}
            \includegraphics[width=\linewidth]{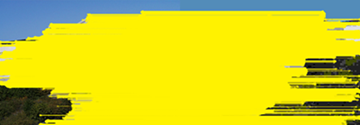}\\[-5pt]
            \caption*{\centering 40,204 Correct Matches}
            \includegraphics[width=\linewidth]{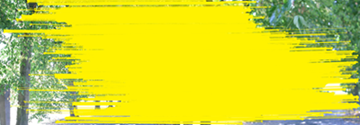}\\[-5pt]
            \caption*{\centering 9,368 Correct Matches}
        \end{minipage}
    \caption{\centering Qualitative comparison of DeepDetect with SIFT under default and extremely low \cite{ref_8} thresholds. Yellow lines show correct correspondences between the images. DeepDetect provides dense and substantially higher correct matches, without requiring manual tuning (due to its intelligent nature).}
    \vspace{-0.3cm}
    \label{fig:fig_6}
\end{figure*}

\begin{figure*}[!t]
	\centering
        \begin{minipage}[t]{0.244\textwidth}
            \centering
            \includegraphics[width=\linewidth]{figures/a1.png}
            \\[3pt]
            \includegraphics[width=\linewidth]{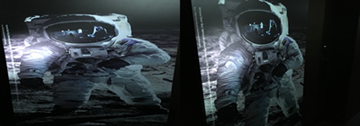}\\[3pt]
            \includegraphics[width=\linewidth]{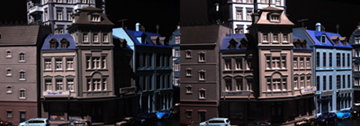}\\[3pt]
            \includegraphics[width=\linewidth]{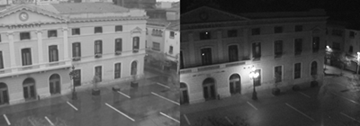}\\[3pt]
            \includegraphics[width=\linewidth]{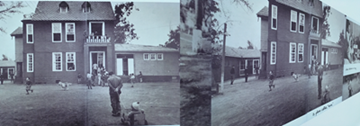}\\[3pt]
            \includegraphics[width=\linewidth]{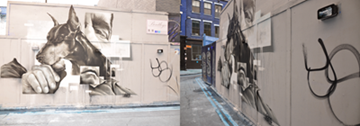}\\[3pt]
            \includegraphics[width=\linewidth]{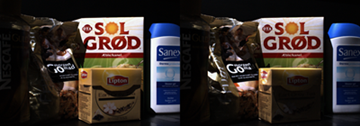}\\[3pt]
            \includegraphics[width=\linewidth]{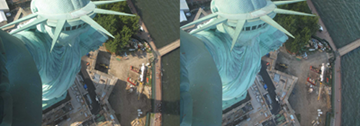}\\[3pt]
            \includegraphics[width=\linewidth]{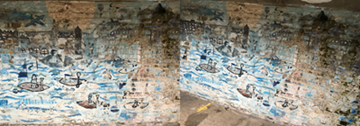}\\[3pt]
        \end{minipage}
        \begin{minipage}[t]{0.244\textwidth}
            \centering
            \includegraphics[width=\linewidth]{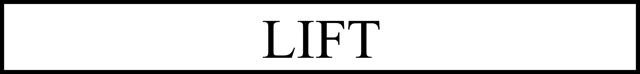}\\[3pt]
            \includegraphics[width=\linewidth]{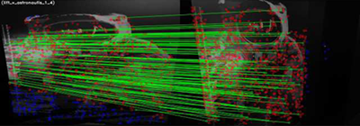}\\[3pt]
            \includegraphics[width=\linewidth]{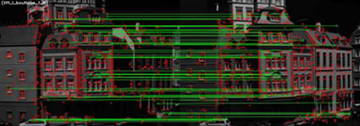}\\[3pt]
            \includegraphics[width=\linewidth]{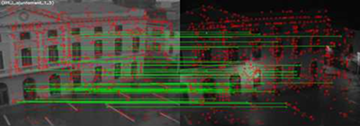}\\[3pt]
            \includegraphics[width=\linewidth]{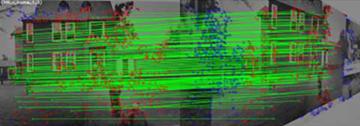}\\[3pt]
            \includegraphics[width=\linewidth]{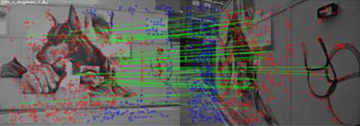}\\[3pt]
            \includegraphics[width=\linewidth]{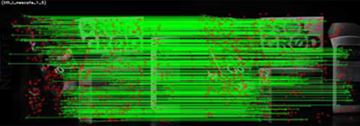}\\[3pt]
            \includegraphics[width=\linewidth]{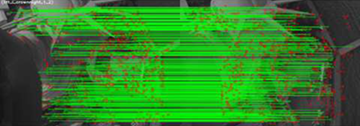}\\[3pt]
            \includegraphics[width=\linewidth]{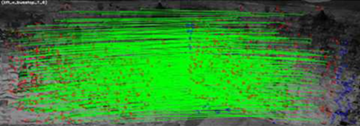}\\[3pt]
        \end{minipage}
        \begin{minipage}[t]{0.244\textwidth}
            \centering
            \includegraphics[width=\linewidth]{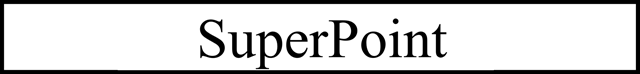}\\[3pt]
            \includegraphics[width=\linewidth]{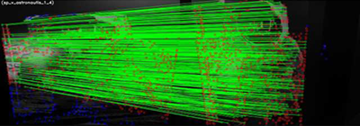}\\[3pt]
            \includegraphics[width=\linewidth]{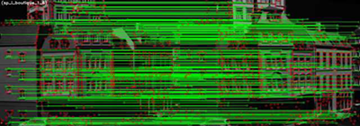}\\[3pt]
            \includegraphics[width=\linewidth]{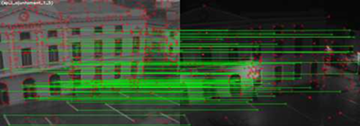}\\[3pt]
            \includegraphics[width=\linewidth]{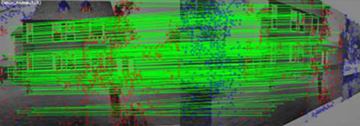}\\[3pt]
            \includegraphics[width=\linewidth]{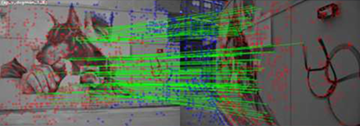}\\[3pt]
            \includegraphics[width=\linewidth]{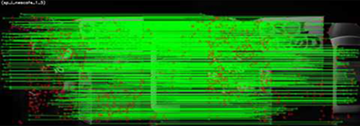}\\[3pt]
            \includegraphics[width=\linewidth]{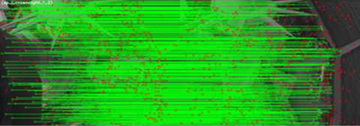}\\[3pt]
            \includegraphics[width=\linewidth]{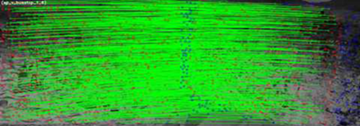}\\[3pt]
        \end{minipage}
        \begin{minipage}[t]{0.244\textwidth}
            \centering
            \includegraphics[width=\linewidth]{figures/a4.png}\\[3pt]
            \includegraphics[width=\linewidth]{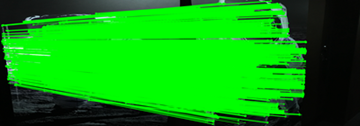}\\[3pt]
            \includegraphics[width=\linewidth]{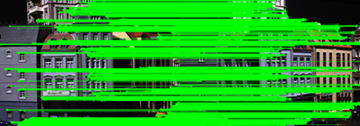}\\[3pt]
            \includegraphics[width=\linewidth]{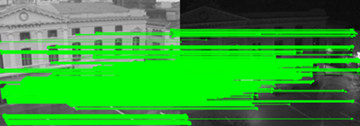}\\[3pt]
            \includegraphics[width=\linewidth]{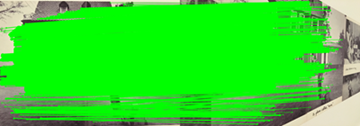}\\[3pt]
            \includegraphics[width=\linewidth]{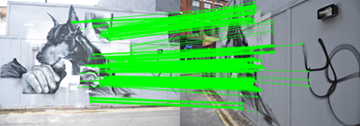}\\[3pt]
            \includegraphics[width=\linewidth]{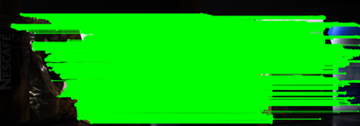}\\[3pt]
            \includegraphics[width=\linewidth]{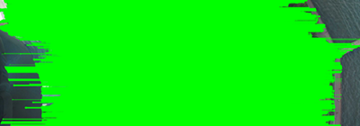}\\[3pt]
            \includegraphics[width=\linewidth]{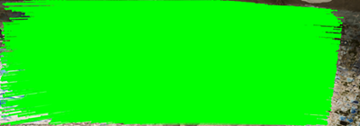}\\[3pt]
        \end{minipage}
    \caption{\centering Qualitative comparison of DeepDetect, LIFT \cite{ref_36}, and SuperPoint \cite{ref_22} on HPatches \cite{ref_37}. Green lines show the correct matches between the image pairs: \textit{(astronautis\_1\_4, boutique\_1\_6, ajuntament\_1\_5, home\_1\_3, dogman\_1\_6, nescafe\_1\_5, crownnight\_1\_2, busstop\_1\_6)}. DeepDetect provides dense and substantially higher correct matches across all scenes, achieving over 0.24 million and 0.33 million matches for the final two image pairs, respectively.
    }
    \vspace{0.2cm}
    \label{fig:fig_7}
\end{figure*}

\begin{table*}[!h]
\definecolor{my_blue}{RGB}{0,0,255}
\renewcommand{\arraystretch}{1.1}
\centering
\scriptsize
\caption{\centering Comparison of Correct Matches Produced by Keypoint Detectors for Selected Image Pairs from Oxford dataset \cite{ref_30}.}
\label{tab:tab_2}
\begin{tabular}{lcccccccc}
\hline
\footnotesize \textbf{Image Pair} & \textbf{SIFT (Low)} & \textbf{ORB (Low)} & \textbf{BRISK (Low)} & \textbf{FAST (Low)} & \textbf{AGAST (Low)} & \textbf{Harris (Low)} & \textbf{Shi-Tomasi (Low)} & \footnotesize \textbf{DeepDetect} \\
\hline
Graf (1--6)    & 20    & 5     & 6     & 13    & 5      & 54    & 6      & \textcolor{my_blue}{\textbf{102}}      \\
Boat (1--2)    & 6517  & 3749  & 3836  & 1703  & 16238  & 33385 & 2164   & \textcolor{my_blue}{\textbf{47447}}   \\
Wall (1--6)    & 66    & 13    & 14    & 2156  & 2357   & 4270  & 175    & \textcolor{my_blue}{\textbf{4756}}    \\
Bikes (1--6)   & 268   & 577   & 358   & \textcolor{my_blue}{\textbf{18112}} & 15458  & 4652  & 449    & 16581   \\
Leuven (1--6)  & 7822  & 3582  & 3163  & 55427 & 53972  & 24296 & 4159   & \textcolor{my_blue}{\textbf{59003}}   \\
Trees (1--6)   & 134   & 234   & 181   & 12078 & \textcolor{my_blue}{\textbf{12185}}  & 3834  & 178    & 9368    \\
UBC (1--6)     & 790   & 977   & 464   & 13024 & 11327  & 28655 & 257    & \textcolor{my_blue}{\textbf{40204}}   \\
\hline
\end{tabular}
\vspace{-0.20cm}
\end{table*}

\textbf{Foreground-Keypoint (F-KP) Ratio:} Evaluates the ability of a detector to detect keypoints within the foreground regions in images. It shows how well a detector focuses on foreground regions rather than the background or irrelevant areas with low or no texture. Let $N_F$ be the number of keypoints in the foreground region and $N_T$ the total detected keypoints, then:
\begin{equation}
\text{Foreground-to-Keypoint Ratio} = \frac{N_F}{N_T}
\label{eq:eq_7}
\end{equation}

\textbf{Correct Matches:} Higher number of correct matches indicates better performance. If $N_A$ and $N_B$ are the number of keypoints detected in two overlapping images $A$ and $B$, then $N_C$ is the number of correct matches between the two sets of keypoints, such that the keypoints of image $A$ are projected onto image $B$ using the ground-truth Homography, and the distance of each keypoint in $A$ to its corresponding keypoint in $B$ is less than a threshold (we used threshold of $1.0$ pixel).

\begin{figure*}[!t]
	\centering
        \begin{minipage}[t]{0.164\textwidth}
            \centering
            \includegraphics[width=\linewidth]{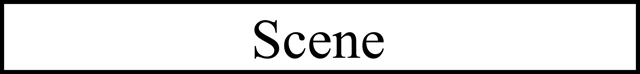}
            \\[3pt]
            \includegraphics[width=\linewidth]{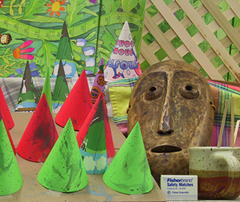}\\[-5pt]
            \caption*{\centering Cones \cite{ref_38}}
            \includegraphics[width=\linewidth]{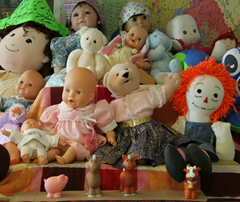}\\[-5pt]
            \caption*{\centering Dolls \cite{ref_39}}
            \includegraphics[width=\linewidth]{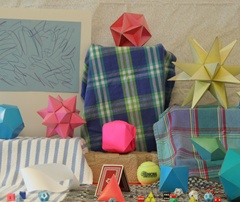}\\[-5pt]
            \caption*{\centering Moebius \cite{ref_39}}
            \includegraphics[width=\linewidth]{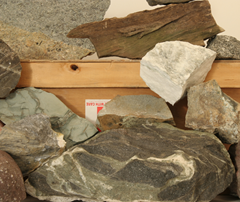}\\[-5pt]
            \caption*{\centering Rocks \cite{ref_39}}
            \includegraphics[width=\linewidth]{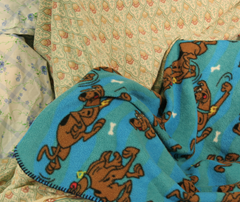}\\[-5pt]
            \caption*{\centering Cloth \cite{ref_39}}
        \end{minipage}
        \hspace{1.1cm}
        \begin{minipage}[t]{0.164\textwidth}
            \centering
            \includegraphics[width=\linewidth]{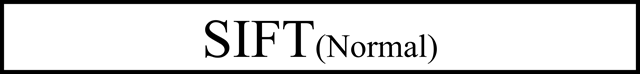}\\[3pt]
            \includegraphics[width=\linewidth]{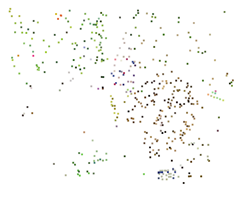}\\[-5pt]
            \caption*{\centering 550 Voxels}
            \includegraphics[width=\linewidth]{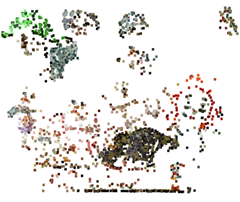}\\[-5pt]
            \caption*{\centering 3,788 Voxels}
            \includegraphics[width=\linewidth]{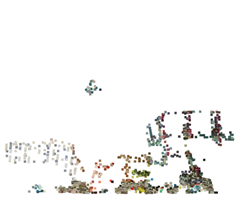}\\[-5pt]
            \caption*{\centering 2,527 Voxels}
            \includegraphics[width=\linewidth]{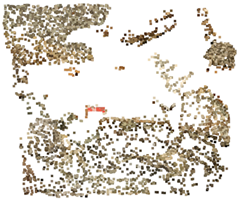}\\[-5pt]
            \caption*{\centering 5,230 Voxels}
            \includegraphics[width=\linewidth]{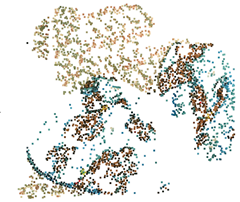}\\[-5pt]
            \caption*{\centering 5,270 Voxels}
        \end{minipage}
        \hspace{1.1cm}
        \begin{minipage}[t]{0.164\textwidth}
            \centering
            \includegraphics[width=\linewidth]{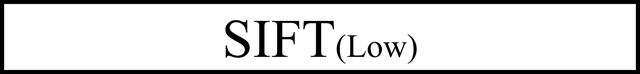}\\[3pt]
            \includegraphics[width=\linewidth]{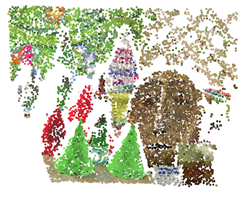}\\[-5pt]
            \caption*{\centering 14,128 Voxels}
            \includegraphics[width=\linewidth]{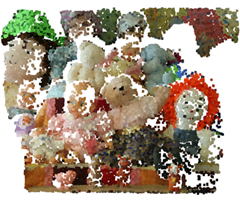}\\[-5pt]
            \caption*{\centering 39,557 Voxels}
            \includegraphics[width=\linewidth]{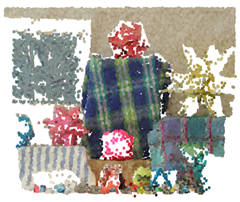}\\[-5pt]
            \caption*{\centering 47,016 Voxels}
            \includegraphics[width=\linewidth]{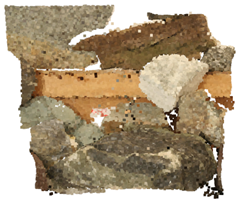}\\[-5pt]
            \caption*{\centering 80,529 Voxels}
            \includegraphics[width=\linewidth]{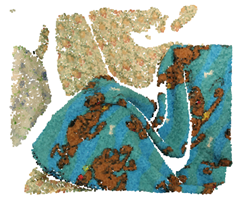}\\[-5pt]
            \caption*{\centering 67,973 Voxels}
        \end{minipage}
        \hspace{1.1cm}
        \begin{minipage}[t]{0.164\textwidth}
            \centering
            \includegraphics[width=\linewidth]{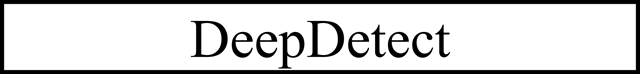}\\[3pt]
            \includegraphics[width=\linewidth]{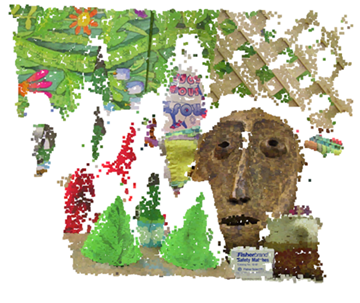}\\[-5pt]
            \caption*{\centering 49,979 Voxels}
            \includegraphics[width=\linewidth]{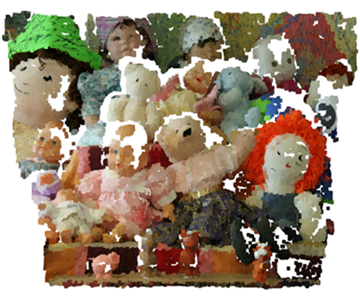}\\[-5pt]
            \caption*{\centering 540,780 Voxels}
            \includegraphics[width=\linewidth]{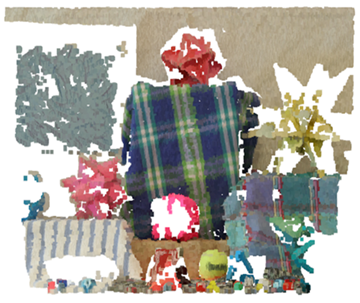}\\[-5pt]
            \caption*{\centering 363,703 Voxels}
            \includegraphics[width=\linewidth]{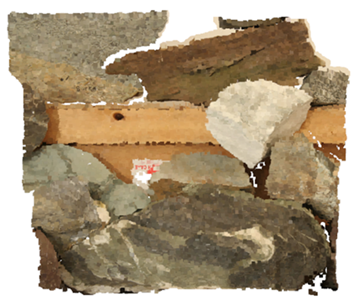}\\[-5pt]
            \caption*{\centering 709,425 Voxels}
            \includegraphics[width=\linewidth]{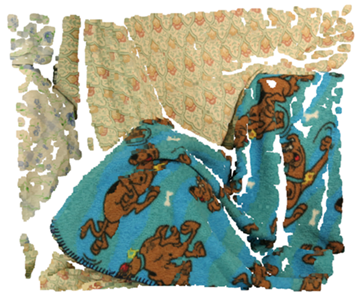}\\[-5pt]
            \caption*{\centering 842,045 Voxels}
            \vspace{0.1cm}
        \end{minipage}
    \caption{\centering Comparison of DeepDetect with SIFT for 3D reconstruction (using only two images per scene) on Middlebury datasets \cite{ref_38,ref_39}. Model used here is trained only on normally illuminated images to avoid low-visibility bias, boosting DeepDetect's performance for scene geometry. Each row (left to right), shows the left camera image of scene, followed by the 3D reconstructed point clouds using SIFT (default thresholds), SIFT (extremely low thresholds) \cite{ref_8}, and DeepDetect. DeepDetect produces highly dense 3D reconstructions, outperforming SIFT even when extremely low thresholds are used.}
    \vspace{-0.3cm}
    \label{fig:fig_8}
\end{figure*}

\begin{table}[!h]
\definecolor{my_blue}{RGB}{0,0,255}
\renewcommand{\arraystretch}{1.26}
\centering
\scriptsize
\vspace{0.18cm}
\caption{\centering Comparison of DeepDetect's repeatability with four well-known learning based detection and description algorithms (MagicPoint \cite{ref_22}, SuperPoint \cite{ref_22}, QuadNet \cite{ref_35}, and R2D2 \cite{ref_23}) on HPatches dataset \cite{ref_37} and Oxford dataset \cite{ref_30}.}
\label{tab:tab_3}
\begin{tabular}{cccc}
\hline
\footnotesize \textbf{HPatches Dataset} & \footnotesize \textbf{MagicPoint} & \footnotesize \textbf{SuperPoint} & \footnotesize \textbf{DeepDetect} \\
\hline
57 Illumination Scenes    & 0.575    & 0.652     & \textcolor{my_blue}{\textbf{0.896}}      \\
59 Viewpoint Scenes    & 0.322	  & 0.503   	& \textcolor{my_blue}{\textbf{0.901}}   \\
\hline
\vspace{0.15cm}
\end{tabular}

\begin{tabular}{cccc}
\hline
\footnotesize \textbf{Oxford Dataset} & \footnotesize \textbf{QuadNet} & \footnotesize \textbf{R2D2} & \footnotesize \textbf{DeepDetect} \\
\hline
Graf    & 0.25    & 0.47     & \textcolor{my_blue}{\textbf{0.89}}      \\
Wall    & 0.46	  & 0.71   	& \textcolor{my_blue}{\textbf{0.98}}   \\
Bark    & 0.16  & 0.47    & \textcolor{my_blue}{\textbf{0.91}}    \\
Boat    & 0.29  & 0.57    & \textcolor{my_blue}{\textbf{0.99}}   \\
Leuven  & 0.77  & 0.77   & \textcolor{my_blue}{\textbf{0.99}}   \\
Bikes   & 0.57 & 0.76   & \textcolor{my_blue}{\textbf{0.94}}   \\
Trees   & 0.50  & 0.60    & \textcolor{my_blue}{\textbf{0.95}}    \\
UBC     & 0.68  & 0.68    & \textcolor{my_blue}{\textbf{0.90}}   \\
\hline
\end{tabular}
\vspace{-0.32cm}
\end{table}

\vspace{0.17cm}
Results show that DeepDetect achieves state-of-the-art performance. On Oxford dataset \cite{ref_30}, it gives an \textbf{average repeatability score} of $\textbf{0.9582}$, surpassing other detectors by a significant margin (see Figure~\ref{fig:figure_1}). The results for \textbf{average keypoint density} are presented in Table~\ref{tab:tab_1}, where DeepDetect achieves the highest average density of $\textbf{0.5143}$. Table~\ref{tab:tab_1} also reports average foreground-to-keypoint ratio on $18$ custom-selected images with ground-truth binary masks. DeepDetect achieves good ratios, indicating that its keypoints are not only dense but also well-concentrated in semantically important regions in the images. With default thresholds, the classical detectors yield higher F-KP ratios but significantly lower keypoint densities. Conversely, reducing thresholds to increase density decreases the F-KP ratio, highlighting the presence of noisy keypoints. For instance, FAST and AGAST with low thresholds, achieve higher densities and lower F-KP ratios, showing their inability to focus salient regions. DeepDetect produces considerably higher densities while maintaining strong foreground focus. Visual comparisons in Figure~\ref{fig:fig_2} highlight how DeepDetect focuses on semantically meaningful regions and provides high keypoint density. Table~\ref{tab:tab_2} and Figure~\ref{fig:fig_6} show that DeepDetect provides substantially higher \textbf{correct matches} and consistently outperforms classical detectors. As far as learning based algorithms are concerned, Table~\ref{tab:tab_3} shows their \textbf{repeatability} comparison with DeepDetect on two different datasets. DeepDetect provides highest repeatabilities due to its intelligent keypoint-edge fusion pipeline. Similarly, Figure~\ref{fig:fig_7} shows its qualitative comparison with LIFT and SuperPoint on HPatches, highlighting that DeepDetect outperforms them with correct matches reaching as high as \textbf{338,118} for the \textit{busstop} image pair. To further validate the benefits of DeepDetect, we compare its performance with SIFT (one of the best algorithms) for 3D reconstruction in Figure~\ref{fig:fig_8}. It is evident that DeepDetect provides high quality, dense 3D reconstructions with number of voxels reaching as high as \textbf{842,045}, substantially surpassing the performance of SIFT even when extremely low thresholds are used. The strength of DeepDetect lies in its ability to combine the geometric robustness and diversity of keypoint and edge detectors, and adaptability to extreme image variations gained through the use of supervision masks generated with \enquote{different thresholds} for normal and challenging scenes. DeepDetect demonstrates stronger generalization across various conditions without manual tuning due to its \textbf{intelligent nature}, making it well suited for visually volatile, changing, or degraded environments.

\section*{Conclusion}
We introduce an intelligent, all-in-one, dense keypoint detector that unifies the strengths of 7 keypoint and 2 edge detectors using deep learning. This keypoint-edge fusion pipeline is called \textbf{DeepDetect}. By leveraging keypoint-edge fusion, we construct diverse supervision masks that allow learning rich, dense, and semantically focused keypoints. Extensive experiments on Oxford, HPatches, and Middlebury datasets demonstrate that DeepDetect outperforms other detectors in repeatability, keypoint density, number of correct matches, and 3D reconstruction quality. DeepDetect can also maintain robustness under extreme image variations depending on the data used for training. Its ability to produce dense, repeatable, and semantically focused keypoints without manual tuning, make it a well suited methodology for image matching, 3D reconstruction, visual SLAM, AR/VR, object tracking, and autonomous navigation under visually volatile conditions.

\bibliographystyle{IEEEtran}
\bibliography{references} 
\end{document}